\documentclass{article}

\usepackage[english]{babel}							
\usepackage[T1]{fontenc} 							
\usepackage[utf8]{inputenc}							
\usepackage{microtype}								
\usepackage{blindtext}
\usepackage{csquotes}                               
\usepackage{paralist}                               
\usepackage{graphicx}                               
\usepackage{subfigure}                              
\usepackage{float}                                  
\graphicspath{{figures/}}
\usepackage{wrapfig}                                
\usepackage[table]{xcolor}                                 
\usepackage{makecell}                               
\usepackage{nicematrix}                             
\usepackage{colortbl}                               
\usepackage{array}                                  
\usepackage{booktabs}                               
\usepackage{multirow}                               
\usepackage{siunitx}                                
\usepackage{hyperref}                               
\usepackage{bookmark}
\usepackage{tabularx}                               

\usepackage{lipsum}                                 
\usepackage{subfiles}                               

\newif\ifblindreview

\usepackage[accepted]{icml2024}

\usepackage{mathtools} 								
\usepackage{amsmath,amsfonts,amssymb, amsthm}

\usepackage[capitalize,noabbrev]{cleveref}

\theoremstyle{plain}

\theoremstyle{definition}

\theoremstyle{remark}

\usepackage[disable,textsize=tiny]{todonotes}

\newif\ifconference
\conferencetrue

\begin{document}

\twocolumn[
\icmltitle{ProtoP-OD: Explainable Object Detection with Prototypical Parts}
\author{Pavlos Rath-Manakidis, Frederik Strothmann, \\Tobias Glasmachers, Laurenz Wiskott}

\begin{icmlauthorlist}
\icmlauthor{Pavlos Rath-Manakidis}{yyy}
\icmlauthor{Frederik Strothmann}{comp}
\icmlauthor{Tobias Glasmachers}{yyy}
\icmlauthor{Laurenz Wiskott}{yyy}
\end{icmlauthorlist}

\icmlaffiliation{yyy}{Faculty of Computer Science, Ruhr University Bochum, Bochum, Germany}
\icmlaffiliation{comp}{sentin GmbH, Bochum, Germany}

\icmlcorrespondingauthor{Pavlos Rath-Manakidis}{pavlos.rath-manakidis@rub.de}

\icmlkeywords{Machine Learning, XAI, Prototype-based Explainability, Object Detection}

\vskip 0.3in
]

\printAffiliations{}

\begin{abstract}
Interpretation and visualization of the behavior of detection transformers tends to highlight the locations in the image that the model attends to, but it provides limited insight into the \emph{semantics} that the model is focusing on.
This paper introduces an extension to detection transformers that constructs prototypical local features and uses them in object detection.
These custom features, which we call prototypical parts, are designed to be mutually exclusive and align with the classifications of the model.
The proposed extension consists of a bottleneck module, the prototype neck, that computes a discretized representation of prototype activations and a new loss term that matches prototypes to object classes.
This setup leads to interpretable representations in the prototype neck, allowing visual inspection of the image content perceived by the model and a better understanding of the model's reliability.
We show experimentally that our method incurs only a limited performance penalty, and we provide examples that demonstrate the quality of the explanations provided by our method, which we argue outweighs the performance penalty.
\end{abstract}

\section{Introduction}

There is a broad literature on neural architectures for object detection (OD) and instance segmentation with the most prominent lines of work revolving around the YOLO architecture~\cite{Redmon_YOLOv1}, R-CNN~\cite{girshick_RCNN, ren_faster_RCNN, he_mask_RCNN} and detection transformers~\cite{carion_DETR, zhu_deformable_DETR, zhang_dino_detr}. The diversity of methods reflects the range of application domains and the diversity of their respective requirements. In many computer vision applications, such as medical image understanding or non-destructive testing, shortcomings of the OD models or legal frameworks require that a domain expert verifies the outputs of the model before decision-making. Therefore, there is a recurring need to assess the confidence of the model and comprehend what decision criteria and intermediary results or representations were used to derive the final detection proposals and to understand their meaning. Such knowledge allows the user to leverage these representations and to recognize and anticipate the model's weaknesses. In this way, the user can better understand the AI, become more vigilant, compensate for the AI's shortcomings, and improve on its results in a structured way.
Nevertheless, most work on explaining the outputs of such models is limited to reporting saliency maps or salient points in the image; see Section~\ref{par:OD_XAI} for a discussion. Such maps are often post-hoc approximations of the model's logic and do not describe what the model considers important in or about the salient areas in the image.

\begin{figure}[h]
    \centering
    \def\svgwidth{\columnwidth}
    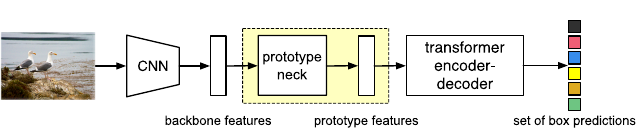
    \caption{\textbf{Detection transformer model with prototype neck.} The prototype neck transforms backbone features into readable prototype maps that are subsequently used for OD. The Figure is adapted from Figure 1 in \citet{carion_DETR}. The yellow block marks our modification to the design. See Figure~\ref{fig:neck_overview} for details. \label{fig:model_overview}}
    \vskip -0.1in
\end{figure}

\begin{figure}[h]
    \centering
    \includegraphics[width=0.9\columnwidth]{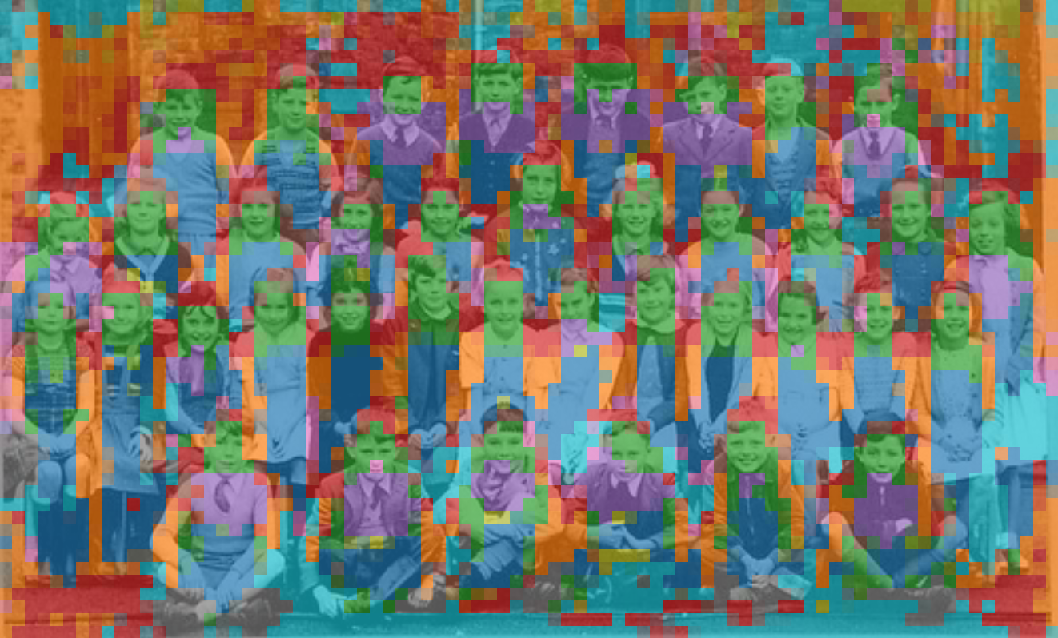}
  \caption{\textbf{A depiction of the most activate prototypes in an image.} Each prototype is represented by a separate color. Most prototypes used to represent the children and the space around them are assigned to the corresponding \texttt{person} class. The prototype in blue focuses on the body of persons, orange on the space between persons, green on heads, and purple on ties. Cyan represents areas where other prototypes dominate. This prototype map has been obtained with the model \texttt{large} described in Section~\ref{sec:experimental_setup}. \label{fig:head_proto_sparse}}
  \vskip -0.1in
\end{figure}

\paragraph{Contribution}
Our contribution to XAI is a modification of detection transformers that allows the model to learn readable intermediate representations in the form of maps of prototypical features, constructed so that only a few features are active at each image location. We name this model variant \emph{ProtoP-OD} (\textbf{Proto}type \textbf{P}art-based \textbf{O}bject \textbf{D}etector). The learned prototypical features are aligned with the known object classes and are mutually exclusive: At each location in the image where there is an object to be detected, only a few prototypes are activated, primarily those assigned to the object's class. These properties allow the user to visually inspect the model's perception of the image using a combination of per-detection attention maps and maps of prototype activations (in short: prototype maps). These two types of maps represent the latent state of the model during decision making. They represent the detection-specific attention and the detection of the prototypes, respectively, for each image location in the model's final representation of the image.

The prototype maps are implemented by inserting a custom neural module into detection transformers between the feature extracting backbone and the transformer head that proposes object detections (see Figure~\ref{fig:model_overview}). This component internally produces sparse maps of prototypical feature activations and propagates only information contained therein to the transformer. We call this component the \emph{prototype neck} and it is a variant of an information bottleneck layer \cite{Koh2020, Schulz2020, tishby_information_2000}. It forces the model to select only the most relevant information in the latent image representation for each image location, resulting in prototypes that encode high-level semantics. We train the prototype neck by rewarding the selection of prototypes that match the target class of the objects to be detected and by quantizing the prototype maps. This yields sets of prototypes that simultaneously reflect the image content relevant for the model, are relatively simple, have distinct semantics, and align with object classes.

\noindent In summary, the core contributions of this paper are the following:

\begin{itemize}
    \item We introduce the prototype neck, a novel module for establishing similarity between data embeddings and prototypes (in the style of \citet{Chen2019_ProtoPNet}) that can produce meaningful representations focusing on just one prototype per embedding, e.g., per image location.
    \item We introduce a novel loss term, the alignment loss, that lets the prototypes in the prototype neck align with the object classes.
    \item We propose ProtoP-OD, a novel prototype-based XAI method for object detection that uses prototypes to represent the areas in the image, causally grounds explanations through an information bottleneck design, aligns prototypes with object classes, and ensures that the prototype activations at each image location are sparse and mutually exclusive.
\end{itemize}

\section{Related Work}
\paragraph{Heatmap-based Explanations for AI Models}
In general, the approaches that visualize the behavior of NNs through maps fall into two categories: saliency maps and prototype maps. Saliency maps highlight the areas that are -- by some definition -- most relevant for the model's output. SHAP~\cite{Lundberg2017_SHAP}, LRP~\cite{Bach2015_LRP}, integrated gradients~\cite{sundararajan_2017_integrated_gradients}, GradCAM~\cite{selvaraju_grad-cam_2017} and methods that display transformer attention patterns \citep[e.g.][]{dosovitskiy_2020_ViT_and_attn_viz, carion_DETR} are some common examples of this category of explanations. Instead, prototype map methods, including our proposal, focus on qualitatively distinct features that emerge in NNs and that highlight image areas with specific semantics. Other prototype map methods are discussed in Section~\ref{subsec:proto_map_XAI_overview}.

\paragraph{Explanations for OD Models} \label{par:OD_XAI}
Work specifically on the explainability of object detector models focuses mostly on producing saliency maps. Most methods adapt GradCAM \cite{kirchknopf_explaining_nodate, dworak_adaptation_2022, yamauchi_spatial_2022, inbaraj_mask_gradcam_2021} to OD, although \citet{kawauchi_shap-based_2022} build on SHAP while \citet{petsiuk_black-box_2021} work with a black-box model and measure the perturbations to its detections that result from masking parts of the input to find salient image areas.
Other visualizations that have been used include plotting the weights of transformer attention mechanisms \cite{carion_DETR, abnar_attn_flow_viz}, visualizing the positions attended to by sparse attention \cite{zhu_deformable_DETR} or the reference points used by certain attention mechanisms \citep[e.g.][]{xia_deformable_attention_transformer}, or visualizing the gradient norm of image input locations with regard to the different outputs of OD \cite{zhu_deformable_DETR}.

In contrast to the saliency- and location-based methods for explaining OD discussed in this section, our method structures the information contained in the image representation of the model in such a way that it is human-readable in its own right.

\section{Methods}
In this section, we describe ProtoP-OD in detail. 
As a base model for OD we use Deformable DETR~\cite{zhu_deformable_DETR} an end-to-end trainable neural network that initially constructs an image representation with dimensions that correspond to the image's height and width, and subsequently employs this representation to propose object detections. ProtoP-OD modifies the architecture by adding one model component, the prototype neck, and the loss function by one loss term, the alignment loss.

\paragraph{Detection Transformers}
Transformers for detection or segmentation (such as DETR~\cite{carion_DETR} (Detection Transformer), Deformable~DETR~\cite{zhu_deformable_DETR}, or SAM~\cite{kirillov_segment_anything_2023}) are models that start by creating an image representation, essentially a tensor of backbone features, from an input image. They then use a transformer~\cite{Vaswani2017_transformer} to process this representation. In this way, a set of output representations are generated that represent the objects identified in the image. The transformer in DETR and Deformable DETR employs an encoder-decoder structure. Initially, in the encoder, the image representation is treated as a set of image location embeddings that pass through multiple layers of multi-head self-attention. Thereby, each embedding gathers information from all others. Subsequently, in the decoder, embeddings, called detection proposals, are introduced and, through several layers of multi-head cross-attention, repeatedly query the image embeddings from the encoder. Each such query involves a selection of image locations to be queried. This can be represented as an attention map. By the end of this process, the detection proposals represent the model's detections.

\paragraph{Notation and Foundational Concepts}
We use the following notation: $B$ is the batch size, $P$ the number of prototypes, $C$ the number of channels, i.e.\ features, that are used in the prototype neck, and $H$ and $W$ denote the height and width of the intermediate image representations, respectively. For each detected object $d$, we calculate an attention map $m_a(d) \in \mathbb{R}^{H \times W}_{\geq 0}$ as the average of the attention maps of all layers and attention heads for this detection in the transformer decoder. ProtoP-OD infers a prototype map $m_p(p) \in \mathbb{R}^{H \times W}_{\geq 0}$ for each image and prototype $p$. It describes the image areas where the prototype is active. A prototype can be any localized class-aligned structure, like a body-part or a delimiter between objects of the same class. To ensure these prototype maps correspond to the object classes, we introduce saliency maps $m_s(d) \in \mathbb{R}^{H \times W}_{\geq 0}$. These maps are used to estimate which areas of the image are relevant for each detection. 

\subsection{Prototype Neck}
\begin{figure}[t]
    \centering
    \includegraphics[width=0.65\columnwidth]{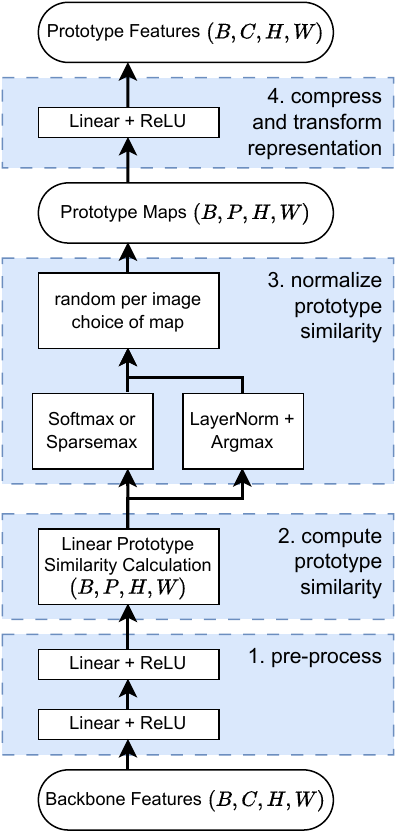}
    \caption{\small \textbf{Structure of the prototype neck.} The image representation from the backbone is processed into features that encode the interpretable prototype maps. Rounded rectangles represent intermediate representations and normal rectangles operations. \label{fig:neck_overview}}
    \vskip -0.1in
\end{figure}

The prototype neck is a neural network module that produces a sparse representation of the image: ideally, only one prototype, i.e. feature, should be active at each image location.
The prototype neck transforms the image representation from the backbone such that it represents the activations of the mutually-exclusive prototypes. Figure~\ref{fig:neck_overview} outlines the structure of the prototype neck. It operates as follows:
\begin{enumerate}
    \item We first process the image representations from the backbone for each image location independently, using two linear layers each followed by ReLU. We do so in order to leverage the pre-trained backbone's generalization ability, by adapting the information contained in its features for the computation of the prototype activations. 
    \item We use a linear layer with bias to obtain similarity scores between the image representations and the prototypes. The latter are encoded as the weights of the linear layer used at this step. That is, we use the inner product to compute similarity.
    \item We form the prototype maps from the prototype similarity scores. We do so either by normalizing the scores with Softmax or Sparsemax~\cite{martins_sparsemax_2016}, depending on the model configuration, or by normalizing them with LayerNorm~\cite{ba_layernorm_2016} to regularize their gradients~\cite{xu_understanding_layernorm_2019} and then apply Argmax. Normalization and the Softmax, Sparsemax, or Argmax operations are performed per image location over the prototypes. This way at each image location the activations of the prototypes form a distribution.
    During training, a subset of the images in each batch is randomly selected and LayerNorm and Argmax are applied to them, while Softmax or Sparsemax are applied to the remaining images.
    \item The sparsely encoded prototype maps are transformed into compact embeddings using a linear layer and ReLU. It is these embeddings that are used by the subsequent components of the OD model.
\end{enumerate}

\paragraph{Sparsemax Neck Variant}
An alternative way to constrain the activations in the prototype maps is to use the Sparsemax function~\cite{martins_sparsemax_2016} instead of Softmax. Sparsemax computes the relative activation of each prototype by linearly projecting the prototype similarity scores onto the probability simplex. This leads to a sparse distribution of prototype activations that is arguably simpler and contains less information, with only a few prototypes active at each location.

\subsection{Alignment Loss} \label{subsec:align-loss}
At image locations that should be relevant for detecting e.g.\ a person, prototypes assigned to the \texttt{person} class should activate. The alignment loss steers the prototype maps towards this behavior.

For each detection $d$, we compute the saliency map $m_s(d)$ over the image locations as a 2-dimensional Gaussian distribution that focuses on the center of the corresponding bounding box by being centered at $\boldsymbol\mu$ and matches its height and width by having covariance $\boldsymbol\Sigma$: 
\[\boldsymbol\mu = \begin{pmatrix} 
 \text{center}_x \\
 \text{center}_y
\end{pmatrix} \text{, } 
\boldsymbol\Sigma = \begin{bmatrix}
\left(\text{width}/6\right)^2 & 0 \\
0 & \left(\text{height}/6\right)^2
\end{bmatrix} \]
We let $m_s(d)$ ignore image locations that correspond to padding around the actual image.

Prior to training, we map each prototype to one of the classes of objects in the dataset.
To compute the alignment loss, we first compute the sum of the probabilities of the prototypes that are assigned to the target class of detection $d$ spatially weighted by $m_s(d)$. This value represents the proportion of aligned prototypes at the area relevant for the detection. This value is offset by $\epsilon = 10^{-3}$ to avoid values close to zero.\footnote{This step is especially useful for quantized prototype activations which can be exactly zero everywhere across the image.} We minimize the negative logarithm of this value averaged over all matched detections:
\[
\mathcal{L}_\text{align} = \frac{1}{D} \sum_{d \in D} -\log\left(\epsilon + m_s(d)^\mathsf{T} \sum_{p \in C(d)} m_p(p)\right) \label{eq:alignment_loss}
\]
Where $D$ is the number of detections and $C(d)$ represents the set of prototypes assigned to the target class of detection $d$.
With $\mathcal{L}_\text{align}$ we introduce an incentive for the class-assigned prototypes to activate near detections of the assigned class and thus specialize to it.

Note, that the alignment loss may not reach zero since multiple objects of different classes can overlap or border on each other, and saliency extends beyond object boundaries.
Additionally, since prototypes are class-aligned and $m_s(d)$ extends into the background, the alignment loss results in background areas being assigned to prototypes of the more common classes.

The alignment loss is used in combination with the classification and the bounding box losses from Deformable DETR to train the model.

\subsection{Winner-takes-all Encoding of Prototype Activations} \label{subsec:argmax_tek}
We want the visible structure of $m_p$ to reflect the most important part of the model's state. To this end, we want the prototype most active at any given image location to be relevant for OD. With winner-takes-all prototype encoding we achieve this by using Argmax to normalize prototype similarity scores: We force subsequent model components to work only with the information of the most active prototypes, because the activation of all other prototypes is set to zero. During training, we choose Argmax for a random subset of the images in each batch. This makes it necessary for the model to be able to perform inference with two different prototype representations: Softmax or Sparsemax and Argmax. Training with the Argmax version of $m_p$ necessitates activating the most informative prototypes the most and forces the model to learn to use this simpler quantized, but more human-readable representation for OD.

Before applying Argmax to the prototype similarity scores, we normalize them with LayerNorm. This normalizes the gradients \cite{xu_understanding_layernorm_2019}. In the backward pass, we propagate the gradients at the quantization stage by using straight-through estimation~(STE)~\cite{gradient_STE_2013}. This means that we skip the Argmax function during the backward pass. Instead, we scale the gradients down by two orders of magnitude. We empirically found that this leads to admissible training behavior.

\section{Experiment Setup and Evaluation Method}

\label{sec:experimental_setup}
We based the implementation for our experiments on Deformable DETR by \citet{zhu_deformable_DETR} and we adapted their design to use a single-scale feature map in order to work with the prototype neck. The base configuration of ProtoP-OD uses a ResNet50~\cite{he_ResNet_2016} backbone, Softmax normalization in the prototype neck, and 300 prototypes.
Model configuration details are given in Appendix~\ref{sec:model_conf_appdx}.

We train and test the models presented in this paper on the COCO~2017~\cite{lin_MS_COCO} dataset. We measure the performance for the configurations we test on the validation split of COCO~2017 (in short: \texttt{val set}) and report the mean and standard deviation over three trials. In the following, we refer to mAP 50\%-95\% over all detections as mAP. Examples of detections, prototype maps, and attention maps in Section~\ref{sec:explainability} come from \texttt{val set}.

Example explanations using our method are obtained on two models: \texttt{large}, which has a ResNet101 backbone and \texttt{sparse\_neck}, which uses the Sparsemax activation function in the prototype neck. Quantitative experiments are performed using the model variants described in Table~\ref{tab:model_variant_configs}.

\begin{table}[h]
    \centering
    \caption{\textbf{Model variants that we evaluated.} We compare performance and explainability scores of these model configurations in Table~\ref{tab:results_overview} and discuss their behavior in Section~\ref{sec:differences_between_model_protos}.}
    \vskip 0.15in
    \begin{small}
    \begin{tabular}[t]{l  l}
    \toprule
        Name & Description \\
        \midrule
        Few Prototypes & \makecell[l]{150 prototypes in prototype neck.} \\
        \midrule
        Sparsemax & Sparsemax prototype normalization. \\
        \midrule
        Argmax & Argmax prototype normalization. \\
        \midrule
        \makecell[l]{Strong \\ Alignment} & \makecell[l]{High prototype alignment loss \\ coefficient value of $8.0$.} \\
    \bottomrule
    \end{tabular}
    \end{small}
    \vskip -0.1in
    \label{tab:model_variant_configs}
\end{table}

\paragraph{Scores for Prototype Map Evaluation} \label{subsec:eval_scores}
To evaluate the explainability of the variants of ProtoP-OD, we propose four scores that characterize the prototype maps' usefulness and clarity. These scores are the exclusion error (EE), measuring the simplicity of prototype maps; the alignment error (AE), assessing the alignment of the prototypes with respect to the target classes; perplexity (PX), evaluating the complexity and diversity of prototype usage; and the average number of active prototypes (AAP), indicating the sparsity of the prototypes. Exact definitions for these scores can be found in Appendix~\ref{sec:eval_scores_details}. For all of the scores, lower values indicate better explainability.

\section{Results} \label{sec:results}
In Section~\ref{sec:explainability} we demonstrate different types of explanations by ProtP-OD. Section~\ref{sec:performance_results} presents the performance-explainability trade-offs we obtained. 

\subsection{Examples of Explanations}
\label{sec:explainability}
The explanations presented in this Section are obtained on models \texttt{large} and \texttt{sparse\_neck} described in Section~\ref{sec:experimental_setup} with performance scores given in Table~\ref{tab:demo_models}. Additional examples of explanations are given in Appendix~\ref{sec:additional_visualisations}.

\begin{table}[h]
    \centering
    \caption{\textbf{Performance of models used for example explanations.}}
    \vskip 0.15in
    \begin{small}
    \begin{NiceTabular}{l | c c}
    \CodeBefore
        \rowcolors{3}{gray!25}{white}
    \Body
        \toprule
        & \multicolumn{2}{c}{Name} \\
        Score & \texttt{large} & \texttt{sparse\_neck} \\
        \midrule
        mAP & 40.69 & 36.73 \\
        PX & 47.82 & 34.14 \\
        AE (\%) & 41.28 & 47.30 \\
        EE (\%) & 59.37 & 62.25 \\
        \makecell[l]{AAP} & 300 & 8.446 \\
        \bottomrule
    \end{NiceTabular}
    \end{small}
    \vskip -0.1in
    \label{tab:demo_models}
\end{table}

\begin{figure}[h]
    \centering
    \def\svgwidth{0.95\columnwidth} 
    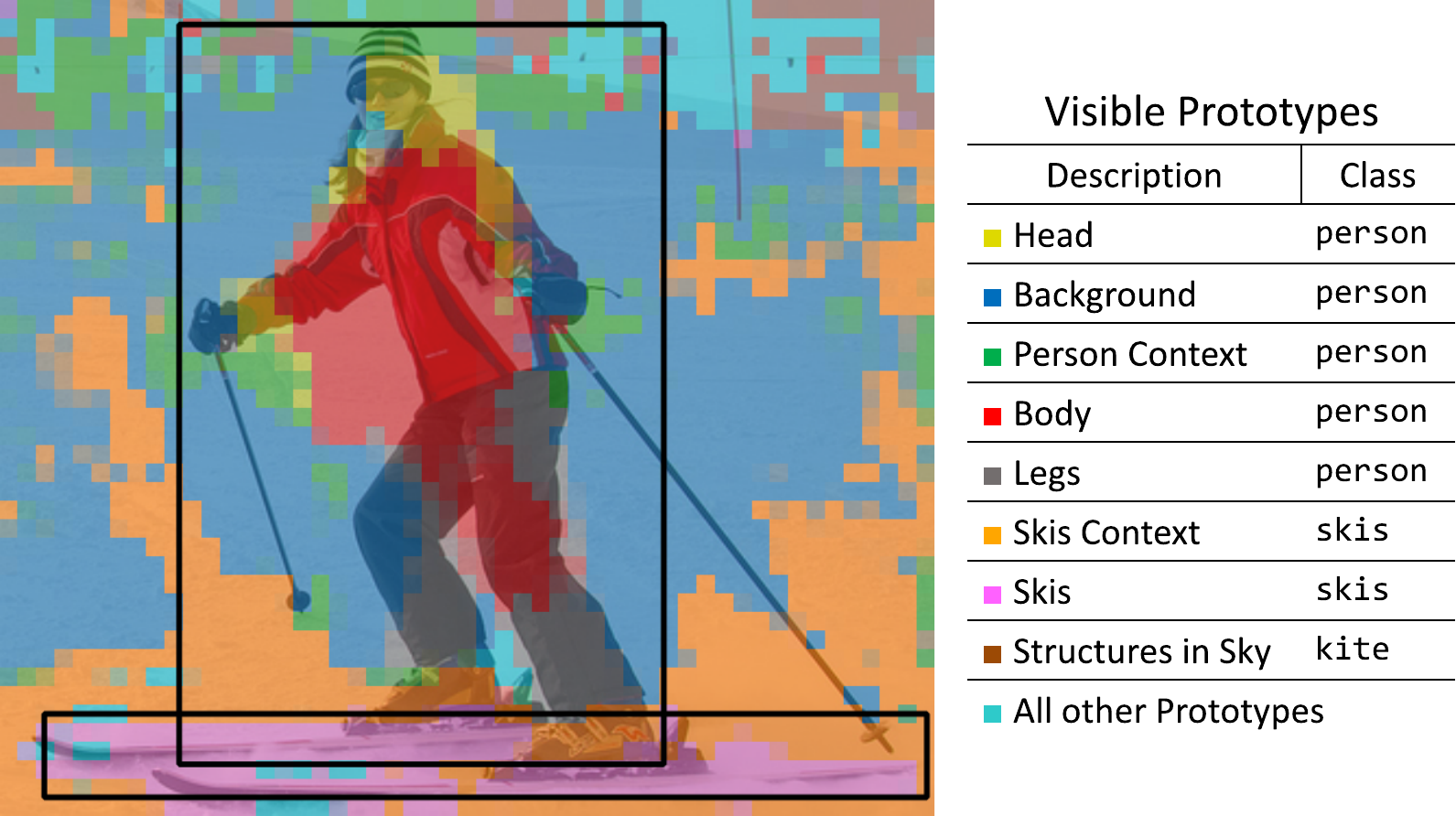
    \caption{\textbf{Multi-prototype map with legend.} Each image location is colored according to the prototypes most active in it, with cyan representing all prototypes that are not colored individually. The semantics of the displayed prototypes also depend on the scale and view of the objects. Background areas are also assigned to prototypes. The map is from model \texttt{large}. \label{fig:multi_proto_showcase}}
    \vskip -0.1in
\end{figure}

\paragraph{Multi-Prototype Maps} \label{sec:multi_prot_map}
It is possible to depict the activations of multiple prototypes in a single map, because the activation of one prototype at an image location suppresses the activation of all other prototypes at that location. Figure \ref{fig:multi_proto_showcase} shows a depiction of
the activations of multiple prototypes, with each prototype represented by a distinct color. To avoid cluttering the image, we only depict the most active prototypes, using an additional color to indicate the activation of all other prototypes.

\begin{figure}[h]
    \centering
    \subfigure[Multi-prototype map. \label{subfig:global_prot_map}]
        {\includegraphics[height=3.1cm]{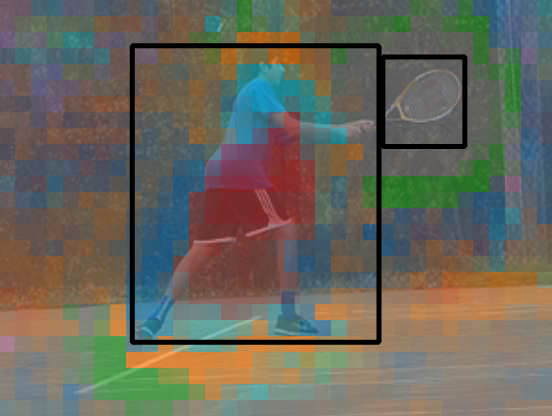}}
    \hfill
    \subfigure[Product map. \label{subfig:product_map}]
        {\includegraphics[height=3.1cm]{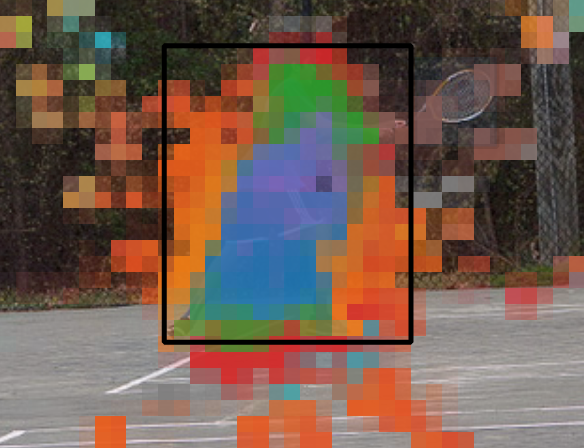}}
  \caption{\textbf{Product map example.} 
  Figure~\ref{subfig:global_prot_map} shows the image-wide multi-prototype map. Some prototypes relevant to the detection of the person are not shown separately. Instead, they are shown in cyan, which represents all prototypes that are not separately colored. 
  Focusing the map on the areas attended to for the detection leads to the product map in Figure~\ref{subfig:product_map}, which shows all relevant prototypes separately.
  The maps are from model \texttt{large}.
  \label{fig:product_map_utility}}
  \vskip -0.1in
\end{figure}

\paragraph{Product Maps} \label{subsec:product_maps}
In order to visualize \emph{where the model attends to what} when it detects an object we combine prototype activations with attention, so that attended image locations are highlighted according to the prototypes active therein. A product map combines the attention map for a single detection $m_a(d)$ and the prototype maps of the prototypes that are most active at the attended locations. The attended prototypes are depicted in distinct colors that are overlaid over the image, in such a manner that they are more visible at the most attended locations.\footnote{Deformable Attention~\cite{zhu_deformable_DETR} is a sparse attention mechanism, resulting in only a few locations being attended to for each detection. To improve readability, we blur the attention maps to make them appear less sparse.} The color intensity through which any given prototype $p$ is shown at each image location is given by the product of the values of $m_a(d)$ and $m_p(p)$ at this location. See Figure~\ref{fig:product_map_utility} for an illustration of how attention can be used to focus on the relevant prototype maps and Figure~\ref{fig:product_maps} for additional examples of product maps.

\subsection{Performance-Explainability Trade-Off}
\label{sec:performance_results}
Table~\ref{tab:results_overview} shows the results of the main experiments. We compared the model variants described in Table~\ref{tab:model_variant_configs}, as we anticipated they would have interesting explainability properties. We describe our observations in Section~\ref{sec:differences_between_model_protos}. We also conducted experiments where we vary the strength of the alignment loss (Table~\ref{tab:alignment_tradeoff}) and the frequency of Argmax prototype activations (Table~\ref{tab:argmax_experiments}) in training. Overall, the experiments establish a reference point for calibrating ProtoP-OD for various explainability requirements. In the tables, the better values among the compared configurations are highlighted with a green background color.

\begin{table*}[h]
    \centering
    \caption{\textbf{Performance and explainability comparison of the model variants from Table~\ref{tab:model_variant_configs}.} }
    \vskip 0.15in
    \begin{small}
    \begin{NiceTabular}{l |  ccccc}
    \CodeBefore
        \rowcolors{3}{gray!25}{white}
    \Body
        \toprule
        & \multicolumn{5}{c}{Model Variant} \\
        Score & \makecell{Few \\ Prototypes} & \makecell{Base \\ Config.} & Sparsemax & Argmax & \makecell{Strong \\ Alignment} \\
        \midrule
        mAP & \colorbox{green!45}{37.8{\scriptsize$\pm$.20}} & \colorbox{green!50}{38.0{\scriptsize$\pm$.11}} & \colorbox{green!25}{36.7{\scriptsize$\pm$.15}} & 33.3{\scriptsize$\pm$.09} & \colorbox{green!5}{35.8{\scriptsize$\pm$.11}} \\
        PX & 51.0{\scriptsize$\pm$7.21} & 71.5{\scriptsize$\pm$17.45} & \colorbox{green!30}{33.4{\scriptsize$\pm$1.62}} & \colorbox{green!50}{22.6{\scriptsize$\pm$1.42}} & 51.0{\scriptsize$\pm$14.61} \\
        \makecell[l]{AE (\%)} & \colorbox{green!5}{20.7{\scriptsize$\pm$.34}} & 20.9{\scriptsize$\pm$.40} & 22.6{\scriptsize$\pm$.91} & 27.9{\scriptsize$\pm$.43} & \colorbox{green!50}{14.4{\scriptsize$\pm$.78}} \\
        \makecell[l]{EE (\%)} & 63.7{\scriptsize$\pm$3.42} & 65.6{\scriptsize$\pm$3.89} & 62.7{\scriptsize$\pm$.90} & \colorbox{green!50}{0.00{\scriptsize$\pm$.00}} & 51.7{\scriptsize$\pm$4.10} \\
        \makecell[l]{AAP} & 150 & 300 & \colorbox{green!30}{8.47{\scriptsize$\pm$.11}} & \colorbox{green!50}{1} & 300 \\
        \bottomrule
    \end{NiceTabular}
    \end{small}
    \vskip -0.1in
    \label{tab:results_overview}
\end{table*}

\begin{table*}[h]
    \centering
    \caption{\textbf{Trade-off between prototype alignment and performance.}}
    \vskip 0.15in
    \begin{small}
    \begin{NiceTabular}{l |  ccccc}
    \CodeBefore
        \rowcolors{3}{gray!25}{white}
    \Body
        \toprule
        & \multicolumn{5}{c}{Alignment Strength} \\
        Score & 0.0 & 0.1 & 0.5 & 2.0 & 8.0 \\
        \midrule
        mAP & \colorbox{green!50}{38.3{\scriptsize$\pm$.17}} & \colorbox{green!50}{38.3{\scriptsize$\pm$.05}} & \colorbox{green!35}{37.9{\scriptsize$\pm$.19}} & \colorbox{green!15}{37.5{\scriptsize$\pm$.17}} & 35.8{\scriptsize$\pm$.11} \\
        PX & 168.{\scriptsize$\pm$25.59} & 140.{\scriptsize$\pm$21.73} & \colorbox{green!25}{77.5{\scriptsize$\pm$16.46}} & \colorbox{green!50}{49.6{\scriptsize$\pm$10.49}} & \colorbox{green!50}{51.0{\scriptsize$\pm$14.61}} \\
        \makecell[l]{AE (\%)} & 96.9{\scriptsize$\pm$.61} & \colorbox{green!15}{45.0{\scriptsize$\pm$1.50}} & \colorbox{green!40}{24.3{\scriptsize$\pm$.45}} & \colorbox{green!45}{17.2{\scriptsize$\pm$.24}} & \colorbox{green!50}{14.4{\scriptsize$\pm$.78}} \\
        \makecell[l]{EE (\%)} & 89.0{\scriptsize$\pm$1.69} & 80.9{\scriptsize$\pm$2.53} & \colorbox{green!5}{67.7{\scriptsize$\pm$6.13}} & \colorbox{green!45}{53.0{\scriptsize$\pm$10.04}} & \colorbox{green!50}{51.7{\scriptsize$\pm$4.10}} \\
        \bottomrule
    \end{NiceTabular}
    \end{small}
    \vskip -0.1in
    \label{tab:alignment_tradeoff}
\end{table*}

\begin{table*}[h]
    \centering
    \caption{\textbf{Trade-off between Argmax prototype quantization and performance.} We vary the frequency at which Argmax is used to compute $m_p$ during training. For the 100\% Argmax configuration Argmax is also used for validation.}
    \vskip 0.15in
    \begin{small}
    \begin{NiceTabular}{l |  ccccc}
    \CodeBefore
        \rowcolors{3}{gray!25}{white}
    \Body
    \toprule
        & \multicolumn{5}{c}{Quantization Frequency} \\
    Score & 0\% & 25\% & 50\% & 75\% & 100\% \\
    \midrule
    mAP & \colorbox{green!50}{38.0{\scriptsize$\pm$.11}} & \colorbox{green!30}{37.1{\scriptsize$\pm$.15}} & \colorbox{green!10}{36.1{\scriptsize$\pm$.21}} & 34.3{\scriptsize$\pm$.20} & 33.3{\scriptsize$\pm$.09} \\
    PX & 80.5{\scriptsize$\pm$7.91} & 69.4{\scriptsize$\pm$4.03} & 54.5{\scriptsize$\pm$3.60} & 74.1{\scriptsize$\pm$20.22} & \colorbox{green!50}{22.6{\scriptsize$\pm$1.42}} \\
    \makecell[l]{AE (\%)} & \colorbox{green!50}{20.1{\scriptsize$\pm$.48}} & \colorbox{green!45}{20.5{\scriptsize$\pm$.75}} & \colorbox{green!20}{22.5{\scriptsize$\pm$.19}} & 25.1{\scriptsize$\pm$.35} & 27.9{\scriptsize$\pm$.43} \\
    \makecell[l]{EE (\%)} & 69.2{\scriptsize$\pm$1.41} & 61.2{\scriptsize$\pm$1.92} & 55.6{\scriptsize$\pm$1.85} & 55.2{\scriptsize$\pm$2.02} & \colorbox{green!50}{0} \\
    \makecell[l]{AAP} & 300 & 300 & 300 & 300 & \colorbox{green!50}{1} \\
    \bottomrule
    \end{NiceTabular}
    \end{small}
    \vskip -0.1in
    \label{tab:argmax_experiments}
\end{table*}

\paragraph{Ablation Study} \label{sec:perf_lit_comp}
We ablate the alignment loss and then also the prototype neck while keeping the architecture and all training and model parameters identical. See Table~\ref{tab:ablations} for the ablation results.
We observe that introducing the prototype neck reduces mAP~50\%-95\% by about 5\% and that introducing the alignment loss reduces the classification loss at the expense of regression performance.
\begin{table}[h]
    \centering
    \caption{\textbf{Ablation study.} CE Loss is the cross-entropy loss for classification and Bbbox Loss and gIoU Loss are the losses for bounding box regression. See \citet{carion_DETR} for the definitions of the losses. \label{tab:ablations}}
    \vskip 0.15in
    \begin{small}
    \begin{NiceTabular}{l |  ccc}
    \CodeBefore
        \rowcolors{3}{gray!25}{white}
    \Body
        \toprule
        Value & Base Config. & \makecell[c]{Alignment Loss \\ Ablation} & \makecell[c]{Neck \\ Ablation} \\
        \midrule
            mAP & 38.0{\scriptsize$\pm$.11} & 38.3{\scriptsize$\pm$.17} & 40.0{\scriptsize$\pm$.18} \\
            CE Loss & 0.392{\scriptsize$\pm$.000} & 0.395{\scriptsize$\pm$.002} & 0.378{\scriptsize$\pm$.000} \\
            Bbox Loss & 0.288{\scriptsize$\pm$.001} & 0.285{\scriptsize$\pm$.001} & 0.274{\scriptsize$\pm$.000} \\
            gIoU Loss & 0.663{\scriptsize$\pm$.002} & 0.657{\scriptsize$\pm$.001} & 0.637{\scriptsize$\pm$.001} \\
        \bottomrule
    \end{NiceTabular}
    \end{small}
    \vskip -0.15in
\end{table}

\section{Discussion of ProtoP-OD as an XAI Method}

\subsection{Plausibility of ProtoP-OD}
ProtoP-OD provides a window into the object detection process. We force the object detector to form class-aligned prototypes and we obtain a legible representation of the internal state of the model that can enable humans using the system to parse the image-input faster and to validate, and potentially interact with, model predictions.

Through the \textbf{alignment loss} we leverage the spatial information of the detections so that image locations where an object of a specific class is depicted get represented with prototypes assigned to that class.

A single \textbf{multi-prototype map} allows for visual inspection of most prototype activations at once. When using Argmax prototype activation quantization or relatively few prototypes, visualizations represent an even larger portion of the prototype information used for OD. To be usable, the information in these maps must be combined with an understanding of the semantics of the prototypes. Prototype semantics can be inferred by observing the activation patterns of the prototypes and identifying representative areas of prototype activation in the data.
Relevant attention information is incorporated into the explanations using \textbf{product maps} to comprehend which areas of prototype activation are important for each detection and to focus the explanation on these areas.

Overall, the explanations by ProtoP-OD are \emph{faithful} and \emph{causally relevant} to how the model operates. Additionally, they are often relatively \emph{legible}. In the case of the Argmax quantized variant of the neck, they can be a \emph{sufficient} description of the information the model preserves for object detection after the backbone.

\paragraph{Limitations}
ProtoP-OD faces challenges with unused and seemingly redundant prototypes. Occasionally, different prototypes may focus on similar or adjacent features or objects, making the explanations inefficient. The model has difficulty distinguishing rare object classes when they frequently appear alongside more common objects, risking misclassification by relying on prototypes of the more common classes. Furthermore, the model's transparency is restricted to the prototype neck, leaving the processes preceding and succeeding this stage opaque. This limitation requires users to experiment to understand how the prototypes behave and how they are used by the subsequent transformer modules.

\subsection{Model Variants} \label{sec:differences_between_model_protos}
In this Section, we discuss the model variants introduced in Table~\ref{tab:model_variant_configs}, the results \ifconference in Table~\ref{tab:results_overview}\fi, and observations about the prototype maps of the model variants on COCO \texttt{val set}. This discussion aims to illustrate how explainability can be configured by configuring ProtoP-OD.

\paragraph{Few-prototypes} 
In this setting, there are either one or two prototypes per class, except for the \texttt{person} class, which has four prototypes available. With this setting, we want to see the effect of using fewer prototypes on model performance and explainability. Perplexity is significantly reduced without a significant reduction in performance. Prototype maps are easier to read, but prototypes tend to activate for multiple different types of image contents. For example, there is a prototype of the class \texttt{person} for both image backgrounds and human torsos. Prototypes are generally less specialized, tend to represent coarser structures, and are spread over larger areas of the image. However, since only a few prototypes are used to represent each image, the prototype maps are easier to read.
\paragraph{Sparsemax} When using Sparsemax, combinations of prototypes are used more extensively to describe image locations than in the base configuration and rototypes tend to have more high-level meaning and clearer contours. With Sparsemax, the activations of most prototypes are exactly zero. This guarantees, that only the locally most active prototypes can influence OD. Some prototypes are not trained and are unused. 
\paragraph{Argmax} Training exclusively with prototype activations quantized by Argmax produces simpler prototype maps but comes with a performance penalty and leads to worse prototype alignment. Otherwise, the prototype maps behave similarly to the base configuration.
\paragraph{Strong Alignment} Increasing the prototype alignment coefficient results in better prototype-class alignment and reduced perplexity (see Table~\ref{tab:alignment_tradeoff}). The prototypes for the \texttt{person} class are more readable and correspond better to parts and views of persons. Model performance in terms on mAP is reduced.

\subsection{Comparison with Related Methods}
In this section we discuss overlap and difference between ProtoP-OD and existing methods.

\subsubsection{Prototype-based Explanations for Classification} \label{subsec:proto_class_XAI}
ProtoP-OD condenses neural network model state in terms of the presence or absence of features that are conceptually abstract and intuitive to humans. In this section we discuss previous work that shares this goal, but for the task of classification.

Methods such as Testing with Concept Activation Vectors~(TCAV) by~\citet{Kim2018_TCAV}, Automated Concept-based Explanation~\cite{Ghorbani2019_ACE}, and Explaining Classifiers with Causal Concept Effect~\cite{goyal_CaCE} attempt to verify and illustrate the importance of user-defined concepts on the inference process of deep classifier models. These approaches calculate the relevance of each prototype for the model without altering the model, i.e., they are post hoc explanation methods. Concept activation vectors~\cite{Kim2018_TCAV} can also be computed per location in latent feature maps.
Our contribution is conceptually similar to using TCAV to create feature activation maps, but unlike TCAV, ProtoP-OD is not a post-hoc technique. Furthermore, TCAV does not aim to represent causally relevant parts of the model state, and the construction of concept activation vectors requires the use of sets of positive and negative examples for each concept. Instead, ProtoP-OD infers prototype features relevant to OD without additional supervision.

\subsubsection{Prototype-Part Models} \label{subsec:proto_map_XAI_overview}
Our notion of prototypes is closest to that defined in the ProtoPNet model by \citet{Chen2019_ProtoPNet}. They define prototypes using similarities between local features from the backbone and learnable vectors representing the prototypes. Thus, the prototypes are locally activated at specific locations in the latent feature maps that correspond to image areas. The prototypes in ProtoPNet are optimized such that in each training image at least one prototype assigned to the target class is maximally active at some location in it and all prototypes from all other classes are suppressed. 

There have been numerous works \cite{Barnett2021_IAIA_LB, Rymarczyk2020_ProtoPShare, Donnelly2021, Hase2019, Nauta2020, li_deep_2018, Rymarczyk2021, sacha_protoseg_2023} that build on this definition of prototypes.

We deviate from ProtoPNet in the design and training process of prototypes for several reasons: \begin{inparaenum}[(i)] \item to simplify the computation of embedding-prototype similarity, \item to treat prototype activations as probabilities, so that they become mutually exclusive, and \item to apply prototype-based reasoning specifically to OD. \end{inparaenum} In ProtoPNet, the prototype activations are computed almost at the end of the model, followed only by a fully connected layer and a non-linearity. We move the prototype computation to the middle of the model, because it is not enough to simply score the aggregate activation of each prototype to perform OD, but we need further modules to process the spatial structure of the prototype activations in order to obtain detection proposals.

The alignment loss we use for explainability also differs from the losses used for explainability in ProtoPNet. In ProtoPNet, the separation cost is introduced as a loss to be minimized to suppress prototypes of non-matching classes in the image. Similarly, in ProtoP-OD, the alignment loss suppresses misaligned prototypes and, because of the distribution-based definition of prototype activations, also fulfills the function that the clustering cost has in ProtoPNet: to attract class-aligned prototypes to detections.

\subsection{Future Directions}
This paper's work can be expanded in several ways.
Firstly, it is crucial to address the performance penalty caused by the prototype neck observed in the experiments in Section~\ref{sec:perf_lit_comp}. We aim to integrate the prototype neck into OD systems without impacting performance while preserving prototype clarity. This is specially important for the version with quantized prototype activations, which allows complete visualization of output-relevant prototype information but faces the largest performance challenges.

In the context of model exploration, debugging, and interaction in general, experiments should determine the extent to which ProtoP-OD relies on individual prototypes. Prototype activations could be removed from images and the model response observed. Prototype pruning and merging could be used to simplify the model and improve explainability, e.g.\ as part of user interaction.

We also envision to integrate ProtoP-OD into an interactive workflow, such that the user can understand and steer the prototypes by interacting with their activations and fine-tuning where certain prototypes should activate.

\section*{Impact Statement}
This paper presents work whose goal is to advance the field of Explainable Artificial Intelligence. There are many potential societal consequences of our work, none which we feel must be specifically highlighted here.

\ifblindreview  \else
\section*{Acknowledgments}
We thank Robin Schiewer, Moritz Lange, and Max Bauroth for insightful discussions and their technical expertise, as well as for the numerous suggestions they provided during this project. We also thank Shirin Reyhanian for her extensive assistance during the literature research.

This work is funded by the German Federal Ministry of Education and Research (BMBF) within the
\enquote{The Future of Value Creation – Research on Production, Services and Work} program (02L19C200), a project that is managed by the Project Management Agency Karlsruhe (PTKA).
\fi

\bibliography{main}
\bibliographystyle{icml2024}

\newpage
\appendix

\section{Model Configuration and Training Details} \label{sec:model_conf_appdx}
The basic configuration of our model consists of a ResNet50 backbone with a dilated C5 stage, followed by the prototype neck with Softmax prototype activations, 300 prototypes, and 256 feature channels. Each class has three to four prototypes assigned to it; except for the \texttt{person} class, which has 7 prototypes, to account for the fact that this class is very heavily represented in the data and is very diverse. We train for 50 epoch. During training some images are normalized with Argmax instead of Softmax. The probability of choosing Argmax increases linearly during training from 0\% to 5\%. Unlike Deformable DETR, the transformer has only two encoder layers. This is because we want the transformer decoder to operate on representations that are similar to the prototype features. The deformable attention in the decoder layers samples 16 points per head.
We schedule the prototype alignment loss coefficient to linearly transition from $1.2$ at the beginning of training to $0.7$ at the final epoch. We use these configurations for all models discussed in the paper unless stated otherwise.

These settings were based on the configuration of Deformable DETR and obtained by trial and error. We could not extensively or systematically optimize the model hyperparameters due to the long training time of our model of approximately $40.5$ hours.

For the model variant with the prototype neck after the encoder and the model ablation without prototype neck, following~\citet{zhu_deformable_DETR}, we place a 1x1 convolution followed by GroupNorm between the backbone and the encoder.

We base the code for our experiments on the public codebase of Deformable DETR (\url{https://github.com/fundamentalvision/Deformable-DETR}) in the version of December 7 2020. We conducted our experiments on a single compute device with 4 NVIDIA A100 GPUs. This device was utilized continuously for around 18 days to conduct all the experiments in this paper.

\section{Definitions of the Scores for Evaluating the Prototype Maps} \label{sec:eval_scores_details}
To evaluate the explainability of our model variants, we propose four values that describe the prototype maps. For all of the scores, lower values indicate better explainability.

The \textbf{Exclusion Error} (EE) quantifies the simplicity of $m_p$ by measuring the extent to which the non-maximally active prototypes are active at each image location averaged across the image:
\[
\text{EE} = \frac{1}{H W} \sum_{i=1}^{H} \sum_{j=1}^{W} \left( 1 - \max_p m_p(p,i,j) \right)
\]
EE ranges from 0 to 1. When reporting experiments, EE is scaled from 0\% to 100\% to enhance readability.

The \textbf{Alignment Error} (AE) measures the percentage of prototype activations that are not aligned with the correct class. It is weighted by the attention each detection proposal directs at each image location and averaged over all detections:
\[
\text{AE} = \frac{1}{D} \sum_{d=1}^{D} \sum_{i=1}^{H} \sum_{j=1}^{W}m_a(d)_{ij} \left( 1 -  \sum_{p \in C(d)} m_p(p,i,j)  \right)
\]
Where $D$ is the number of detections and $C(d)$ represents the set of prototypes assigned to the target class of detection $d$. AE ranges from 0 to 1. When reporting experiments, AE is scaled from 0\% to 100\% to enhance readability.

Adapted from VQ\nobreakdash-VAE~\cite{van_den_oord_2017_VQ-VAE}, \textbf{Perplexity} (PX) measures the complexity of the prototype maps and the diversity of the prototypes used by the model across an entire image. It is the exponentiated entropy of the average of the activation distributions over all locations in $m_p$:
\[
\text{PX} = \exp\left(-\sum_{p=1}^{P} \bar{m}_p \log \bar{m}_p \right)
\]
Where $\bar{m}_p = \frac{1}{H W} \sum_{i=1}^{H} \sum_{j=1}^{W} m_p(p,i,j)$. 

When considering the Sparsemax variant of the neck, it becomes meaningful to count the \textbf{average number of active prototypes} (AAP) per image location to measure the complexity of the prototype representations:
\[
\text{AAP} = \frac{1}{H W} \sum_{i=1}^{H} \sum_{j=1}^{W} \sum_{p=1}^{P} \mathbf{1}\left\{m_p(p,i,j) > 0\right\}
\]
Where $\mathbf{1}\left\{ \cdot \right\}$ is the indicator function.

\section{Additional Examples of Explanations} \label{sec:additional_visualisations}
\begin{figure}[h]
    \centering
    \subfigure[A prototype-feature that detects heads and additionally encodes information about the orientation of the person in the form of the position of the person's neck or back. From the model \texttt{large}.]
    {\includegraphics[height=3cm]{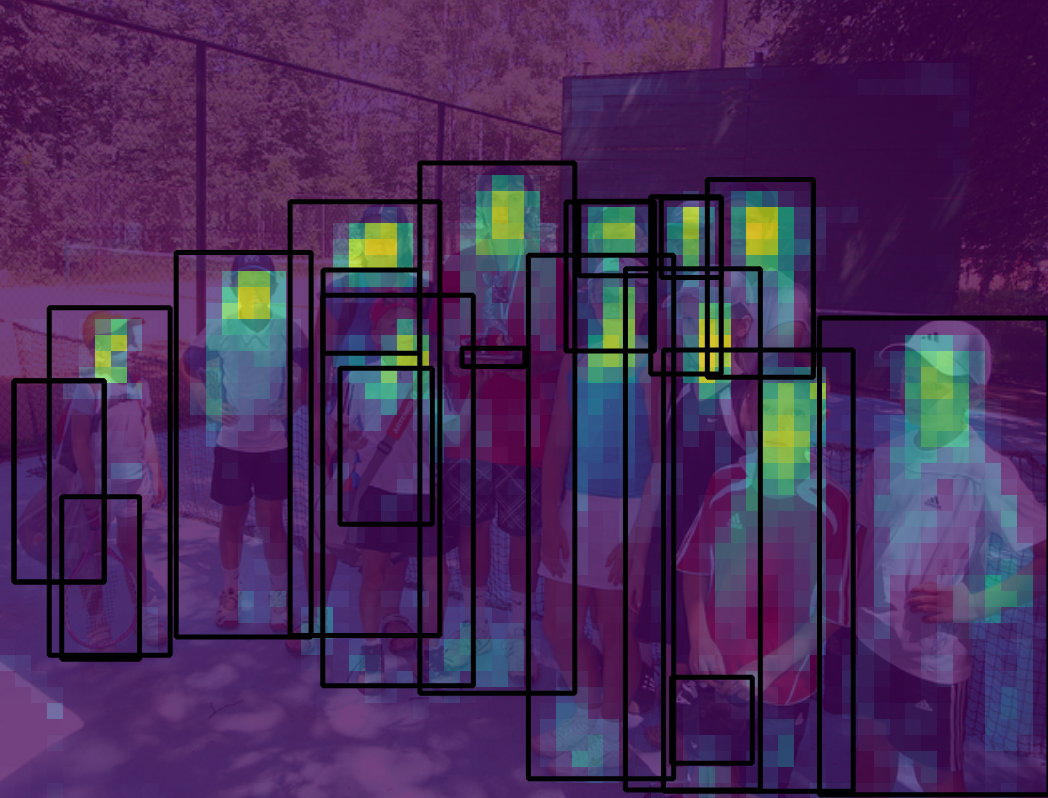}
    \hfill
    \includegraphics[height=3cm]{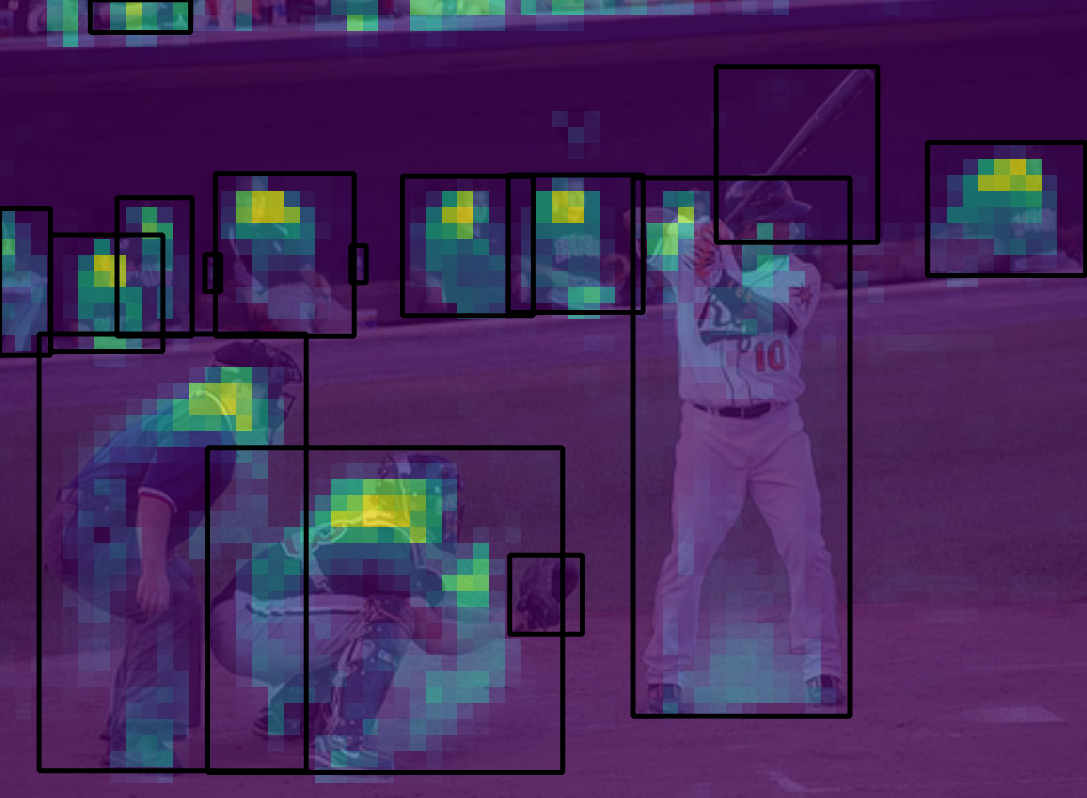} \label{fig:head_proto}}
    
    \subfigure[Prototype marking the area of donuts. Note that between objects the activation of the prototype decreases even if the donuts overlap. From the model \texttt{sparse\_neck}.]
    {\includegraphics[height=2.15cm]{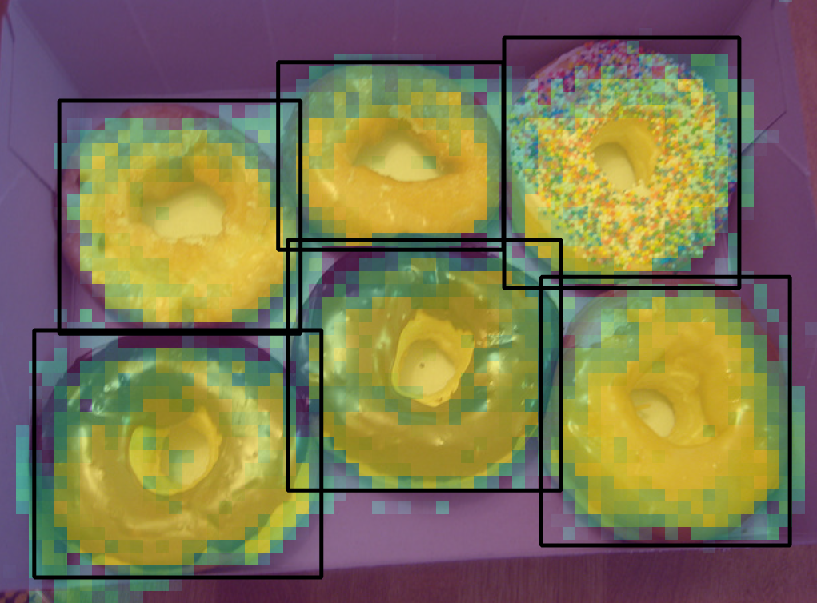}
    \hfill
    \includegraphics[height=2.15cm]{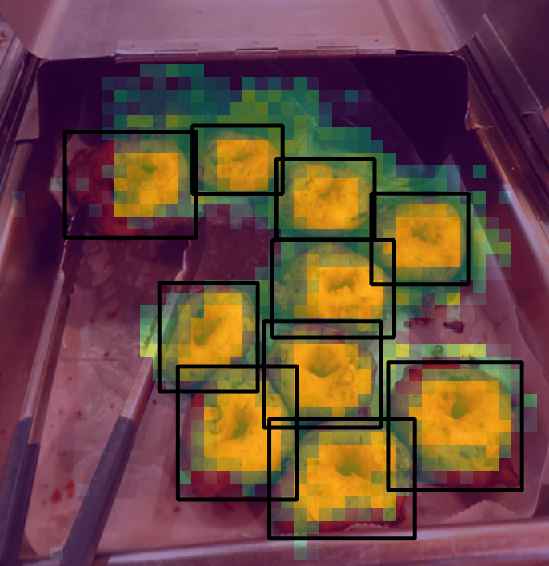}
    \hfill
    \includegraphics[height=2.15cm]{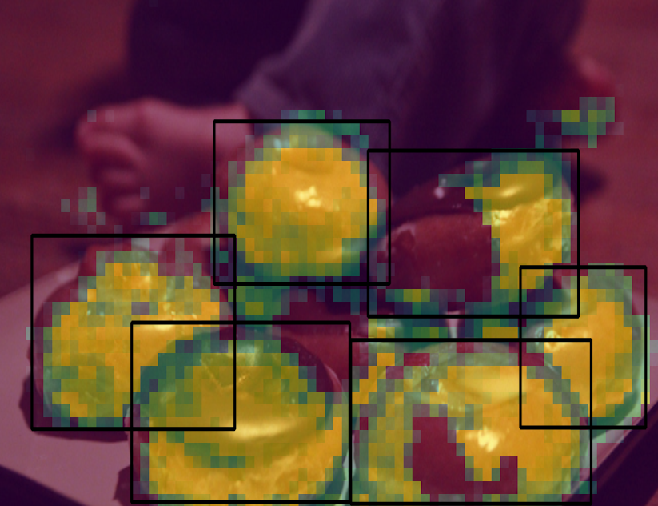}}
    
  \caption{\textbf{Examples of single prototype maps}. Visualizations of the activation of a single prototype. Bright yellow overlay color means that the prototype accounts for 100\% of the total prototype activation at that location and dark blue represents 0\% contribution by the prototype. \label{fig:single_proto_examples}}
\end{figure}

\paragraph{Single Prototype Maps}
Examples of prototype maps that show a single prototype are given in Figure~\ref{fig:single_proto_examples}. Single prototype maps are, in effect, an overlay over the image such that for each image location we can see the extent to which the selected prototype contributes to the model's representation of that location.

\begin{figure}[h]
    \centering
    \subfigure[Close-up of a bookcase. The prototypes assigned to the class \texttt{book} are shown in green, orange, blue, and red. All other prototypes are depicted in purple. Orange denotes books that stand upright, green identifies the bookcase context, blue space above and below books, and red correlates with books that lie on the side. \label{subfig:bookcase}]
    {\includegraphics[height=7cm]{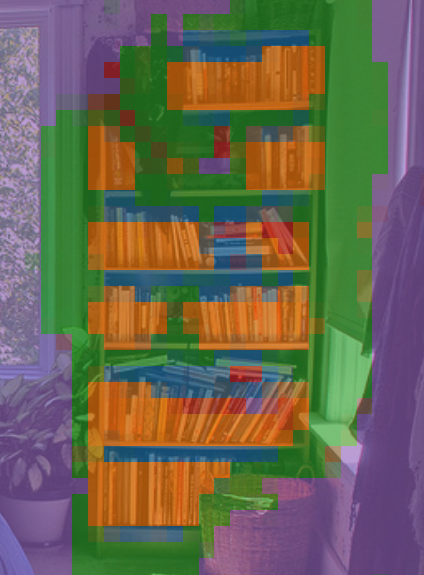}}
    \hfill
    \subfigure[Overlapping teddy bears. The prototypes in blue, orange, and green are assigned to the \texttt{teddy bear} class. The orange and green prototypes focus on edges relevant for separating the teddy bears. \label{subfig:teddy_bears}]
    {\includegraphics[height=7cm]{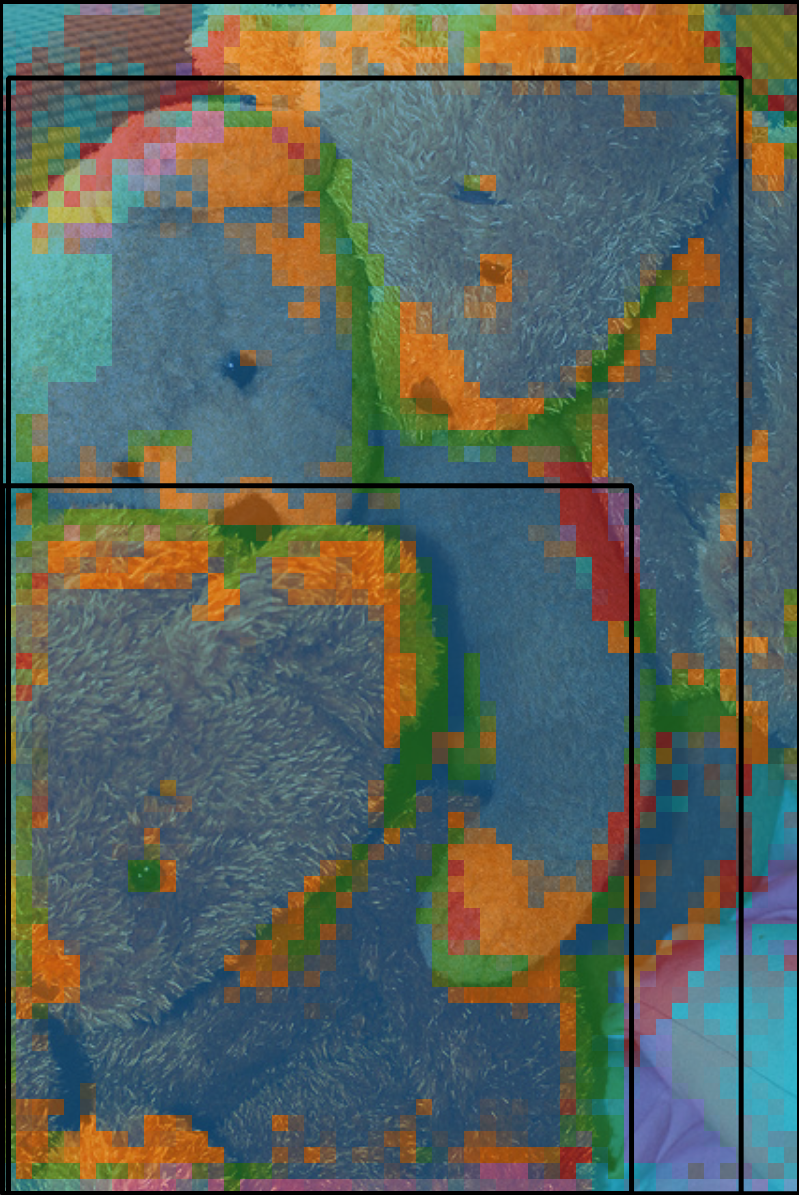}}
  \caption{\textbf{Examples of multi-prototype maps 1.}
  All maps are from model \texttt{large}. \label{fig:multi_proto_examples}}
\end{figure}

\begin{figure}[h]
    \centering
    \subfigure[Apples in a bowl. The prototype in blue focuses on the apples themselves, while other prototypes act as delimiters. All prototypes not depicted separately are depicted in cyan. \label{subfig:apples}]
    {\includegraphics[width=0.95\columnwidth]{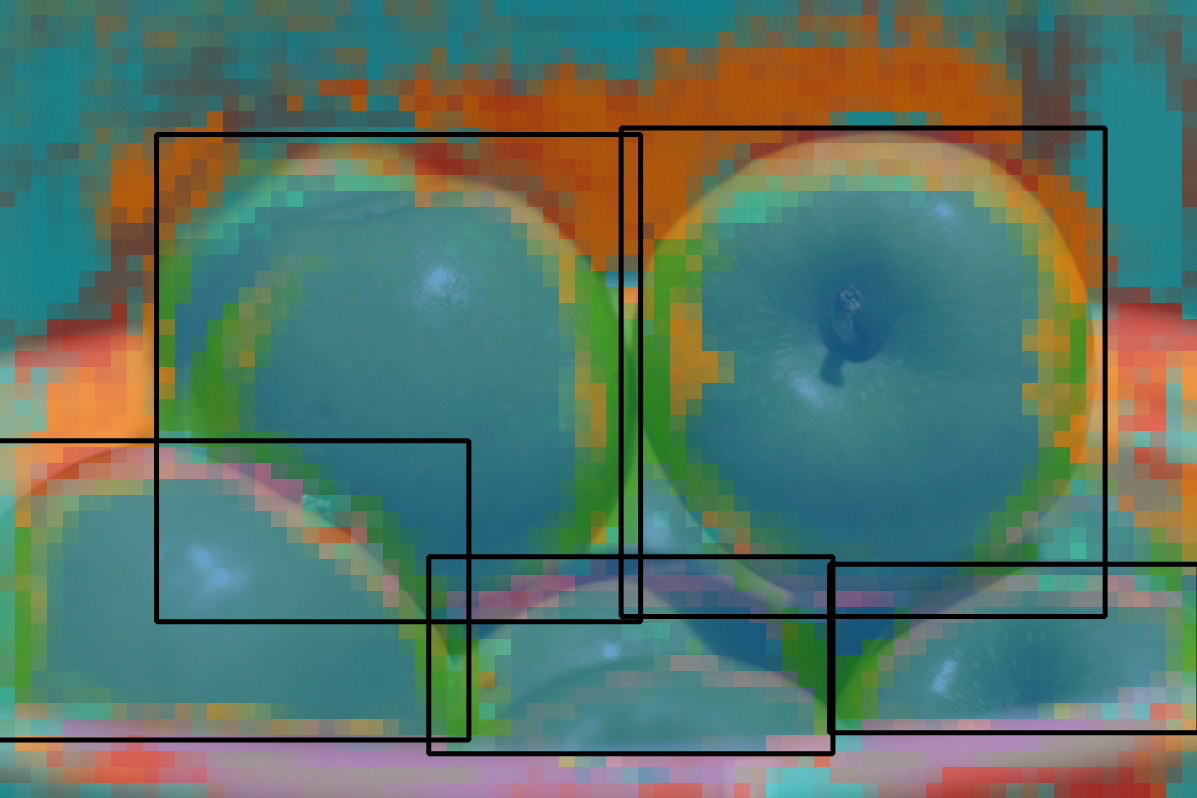}}
    \hfill
    \subfigure[Prototypes assigned to the class \texttt{bird}. The green prototype detects the location of birds and the orange prototype marks context around birds. All prototypes assigned to other classes are depicted in red. \label{subfig:birds}]
    {\includegraphics[width=0.95\columnwidth]{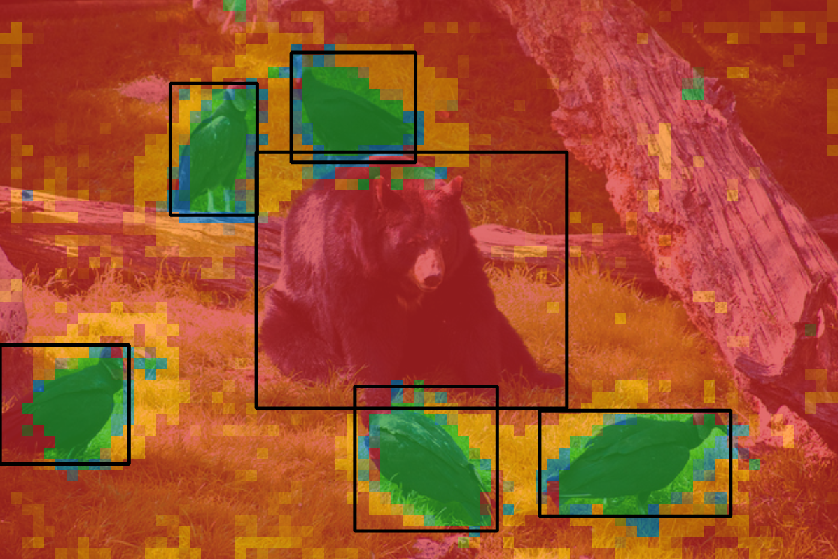}}
  \caption{\textbf{Examples of multi-prototype maps 2.}
  All maps are from model \texttt{large}. \label{fig:multi_proto_examples2}}
\end{figure}

\begin{figure}[h]
    \centering
    \vspace{0.2cm}
    \includegraphics[width=0.75\columnwidth]{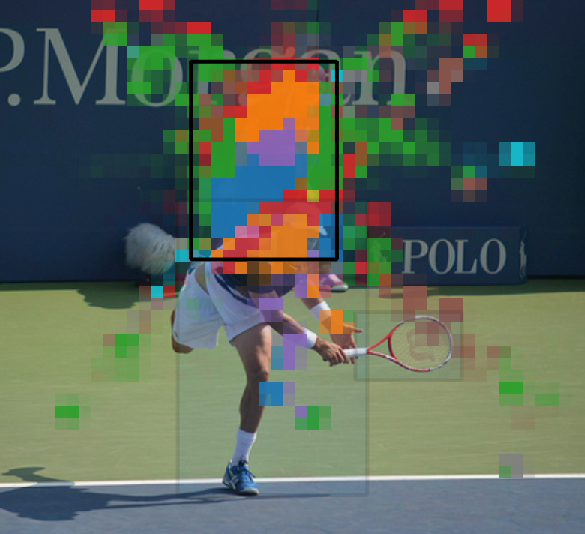}
    \smallskip
    \vspace{0.2cm}
    \includegraphics[width=0.75\columnwidth]{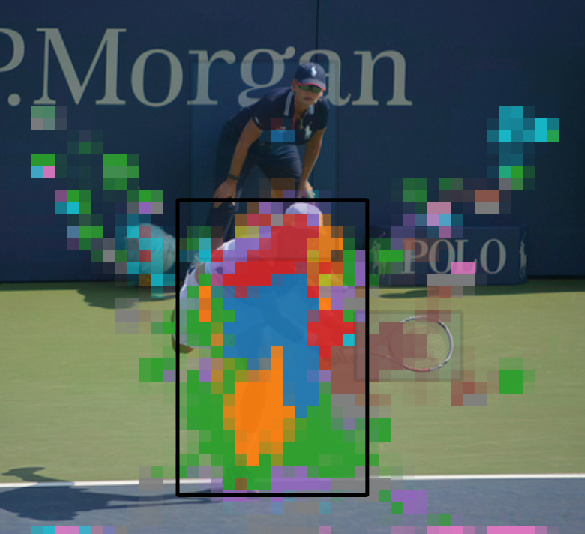}
    \smallskip
    \vspace{0.2cm}
    \includegraphics[width=0.75\columnwidth]{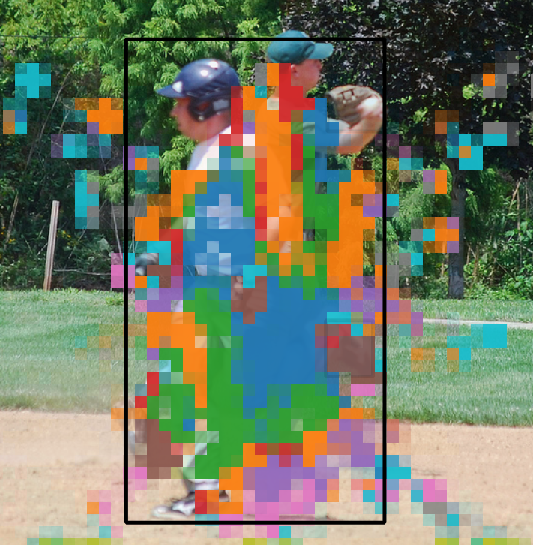}
  \caption{\textbf{Additional examples of product maps of persons.} Different prototypes activate at different parts of the depicted persons and some prototypes are used to separate and delimit them. The assignment of colors to prototypes is not consistent across the images, but is based on the relative activation of the prototypes within the attended areas. The maps are from model \texttt{large}. \label{fig:product_maps}}
\end{figure}

\end{document}